%% file: acl_latex.tex
\pdfoutput=1

\documentclass[11pt]{article}

\usepackage[preprint]{acl}

\usepackage{times}
\usepackage{latexsym}

\usepackage[T1]{fontenc}

\usepackage[utf8]{inputenc}

\usepackage{microtype}

\usepackage{inconsolata}

\usepackage{graphicx}

\usepackage{booktabs}
\usepackage{multicol}
\usepackage{multirow}
\usepackage{hyperref}
\usepackage{colortbl}
\usepackage{array}
\usepackage{amsmath}
\usepackage{caption}
\usepackage{pifont}
\usepackage{amssymb}
\usepackage{lipsum}
\usepackage{paralist}
\usepackage{bbding}

\newcommand{\minisection}[1]{\noindent \textbf{#1}\quad}
\definecolor{reproduce1}{RGB}{0,0,255}
\definecolor{reproduce2}{RGB}{0,100,0}
\definecolor{reproduce3}{RGB}{200,0,0}

\title{Multimodal Machine Translation with Visual Scene Graph Pruning}

\author{
Chenyu Lu\textsuperscript{\rm 1}  \, 
Shiliang Sun\textsuperscript{\rm 2}\Thanks{ Corresponding author} \, 
Jing Zhao\textsuperscript{\rm 1}  \, 
Nan Zhang\textsuperscript{\rm 3}  \, 
Tengfei Song\textsuperscript{\rm 4} \,
Hao Yang\textsuperscript{\rm 4}\\
\textsuperscript{\rm 1} East China Normal University \, 
\textsuperscript{\rm 2} Shanghai Jiao Tong University \, \\
\textsuperscript{\rm 3} Wenzhou University \, 
\textsuperscript{\rm 4} Huawei Technologies Ltd.
}

\begin{document}
\maketitle
\input{sections/0_abstract.tex}
\input{sections/1_introduction.tex}
\input{sections/2_related_work.tex}
\input{sections/3_approach.tex}
\input{sections/4_experiments.tex}
\input{sections/5_conclusion.tex}
\input{sections/6_limitation.tex}

\bibliography{custom}

\appendix
\input{sections/7_appendix.tex}
\label{sec:appendix}

\end{document}

%% file: sections/0_abstract.tex
\begin{abstract}
    Multimodal machine translation (MMT) seeks to address the challenges posed by linguistic polysemy and ambiguity in translation tasks by incorporating visual information. A key bottleneck in current MMT research is the effective utilization of visual data. Previous approaches have focused on extracting global or region-level image features and using attention or gating mechanisms for multimodal information fusion. However, these methods have not adequately tackled the issue of visual information redundancy in MMT, nor have they proposed effective solutions. In this paper, we introduce a novel approach--multimodal machine translation with visual Scene Graph Pruning (PSG), which leverages language scene graph information to guide the pruning of redundant nodes in visual scene graphs, thereby reducing noise in downstream translation tasks. Through extensive comparative experiments with state-of-the-art methods and ablation studies, we demonstrate the effectiveness of the PSG model. Our results also highlight the promising potential of visual information pruning in advancing the field of MMT.
\end{abstract}

%% file: sections/1_introduction.tex
\section{Introduction}
\label{sec:introduction}
The Multimodal Machine Translation (MMT) task~\cite{caglayan-etal-2016-multimodality,elliott-etal-2016-multi30k} aims to enhance traditional neural machine translation by integrating visual information from images to help disambiguate words and phrases that are polysemous or ambiguous. For instance, the word ``crane'' can refer to either a bird or a piece of machinery, and the visual context provided by images can help clarify the intended meaning. By bridging the gap between the visual and language modalities, MMT has the potential to significantly improve translation accuracy and reliability, offering exciting applications across diverse domains.

With the maturation of neural machine translation backbones, effectively utilizing visual modality information and enhancing text-image fusion have emerged as critical bottlenecks in improving the performance of MMT. Early approaches in MMT incorporate visual data through global image features extracted from pretrained CNNs~\cite{calixto-etal-2016-dcu,calixto-liu-2017-incorporating,li2021multi,libovicky-etal-2016-cuni}. While computationally efficient, these methods compress the semantic content of the entire image into a single global feature vector, resulting in substantial information loss that negatively impacts the quality of translation. To address this, more recent studies have focused on extracting region-level or grid-level image features~\cite{li2021multi, zhao2021word, li-etal-2022-vision} and enhancing textual representations through attention or gating mechanisms that incorporate visual information~\cite{li-etal-2022-vision, yin-etal-2020-novel, zhang2020neural, huang-etal-2016-attention, calixto-etal-2017-doubly, tayir2024encoder, zuo-etal-2023-incorporating}. These methods have demonstrated improved performance by selectively aligning visual cues with textual representations.

\input{tables/entities.tex}
However, despite these advancements, a critical aspect of MMT remains largely overlooked: the issue of redundant visual information. In MMT, the principle of ``faithfulness''—one of the key elements in the translation trifecta of ``faithfulness, expressiveness, and elegance''—is the most critical criterion for assessing translation quality. This principle emphasizes that the model should prioritize textual input for translation, using the image modality primarily to provide contextual clues for resolving ambiguities. However, images naturally contain a wealth of information, often surpassing the richness of textual content, with visual entities far outnumbering textual ones. This surplus of image information can undermine translation quality, leading to deviations from the core principle of ``faithfulness''.

To validate this hypothesis, we analyze the average number of entities in the Multi30K dataset. As shown in Table~\ref{tab:entities}, the problem of redundant image information becomes apparent from two key perspectives: (\textit{i}) previous studies commonly rely on pretrained object detection networks to extract $36$ visual entities, which we found to be excessive. Our analysis shows that the average number of reliable visual entities (defined as those with confidence scores above $0.3$) is only $9.17$. Including $36$ entities in downstream tasks, therefore, introduces a substantial amount of noisy and unreliable information; (\textit{ii}) the average number of language entities per sample for English, German, and French corpora is approximately $3.69$, which is far fewer than the number of visual entities, further underscoring the issue of visual information redundancy.

\input{figures/prune.tex}

To mitigate the effects of redundant visual information, we propose a visual Scene Graph Pruning model (PSG) for MMT. As illustrated in Figure~\ref{fig:prune}, the model separately extracts visual and language scene graphs to enhance semantic understanding. The language scene graph is then utilized to guide the pruning of the visual scene graph. Compared to directly using text sequences for pruning, leveraging language scene graphs significantly reduces the heterogeneity gap between visual and language modalities. This approach effectively retains visual information relevant to the text while minimizing the impact of excessive visual noise on downstream machine translation performance. Additionally, we implement a multi-step pruning strategy to prevent excessive loss of critical information during the pruning process.

Our contributions are summarized as follows:	
\begin{compactitem}
	\item We are the first to identify the issue of redundant visual information in MMT and propose a novel visual Scene Graph Pruning (PSG) model, which eliminates unnecessary visual data while preserving text-relevant information.
    \item We simultaneously generate visual and language scene graphs to guide the pruning process, effectively bridging the structural heterogeneity between visual and language modalities and enhancing pruning performance.
	\item We conduct comprehensive comparative experiments and ablation studies on the multilingual datasets Multi30K, AmbigCaps, and CoMMuTE, demonstrating the superiority of PSG.
\end{compactitem}

%% file: tables/entities.tex
\begin{table}[t]
    \centering
    \setlength\tabcolsep{5pt}
    \resizebox{1\linewidth}{!}{
    \begin{tabular}{c|c|ccc}
        \toprule

        \textbf{Previous Work} & \multicolumn{4}{c}{\textbf{Our Preprocess Analysis}}\\
        \midrule
        \multirow{2}{*}{\textbf{\#Visual Entities}} & \multirow{2}{*}
        {\textbf{\#Visual Entities}} &\multicolumn{3}{c}{\textbf{\#Language Entities}} \\
        &&\textbf{English} &\textbf{German} &\textbf{French}\\
        \midrule
        \midrule
        $36.00$ & $9.06$ & $3.48$ & $3.66$ & $3.92$\\
        \bottomrule
    \end{tabular}}
    \caption{Entity number statistic on Multi30K dataset.}
    \label{tab:entities}
\end{table}

%% file: figures/prune.tex
\begin{figure}
    \includegraphics[width=1\linewidth]{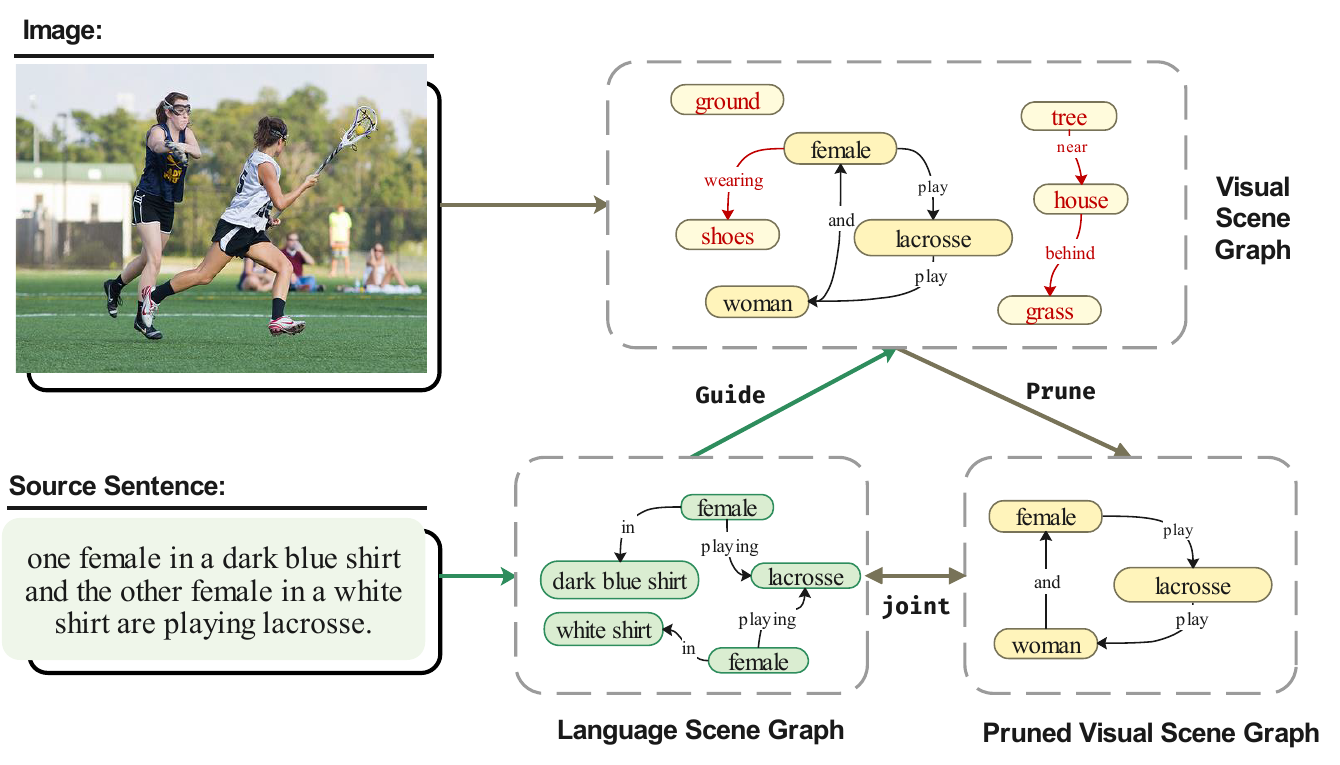}
    \caption{Visual scene graph pruning with guidance from language scene graphs.}
    \label{fig:prune}
\end{figure}

%% file: sections/2_related_work.tex
\section{Related Work}
Multimodal Machine Translation (MMT) seeks to address ambiguities in textual input by integrating information from the visual modality. Early approaches to MMT incorporate global image features extracted from pretrained CNNs~\cite{calixto-etal-2016-dcu,calixto-liu-2017-incorporating,li2021multi,libovicky-etal-2016-cuni}. However, encoding the semantics of an entire image into a single global feature often leads to significant information loss, negatively impacting translation quality. To mitigate this limitation, recent research has shifted toward extracting region-level or grid-level image features~\cite{li2021multi,zhao2021word,li-etal-2022-vision} and employing gating mechanisms to enhance textual representations with visual information~\cite{li-etal-2022-vision,yin-etal-2020-novel,zhang2020neural}, or attention mechanisms to incorporate relevant visual information~\cite{huang-etal-2016-attention,calixto-etal-2017-doubly,tayir2024encoder,zuo-etal-2023-incorporating}, achieving notable performance improvements.

Building on prior work~\cite{fei-etal-2023-scene}, this paper leverages scene graphs~\cite{wang-etal-2018-scene,johnson2015image,yang2019auto,fei-etal-2023-scene} to represent visual information, capturing intricate relations between fine-grained entities to enhance the quality of inputs for translation models. Unlike~\citet{fei-etal-2023-scene}, which addresses the absence of images during inference by reconstructing visual scene graphs via cross-modal mechanisms, our approach directly tackles the challenge of image information redundancy in MMT, ensuring more efficient utilization of visual data.

%% file: sections/3_approach.tex
\input{figures/overview.tex}
\section{Approach}
In this section, we begin by formally defining the Multimodal Machine Translation (MMT) task. Given a source sentence $S$ in a source language $\mathcal{S}$, a corresponding target sentence $T$ in a target language $\mathcal{T}$, and an associated image $I$, the objective is to find a translation $T^*$ such that $T^* \in \mathcal{T}(S)$, where $\mathcal{T}(S)$ represents the set of all possible translations of $S$.

As illustrated in the Figure~\ref{fig:overview}, our proposed PSG model adopts an encoder-decoder architecture to address the MMT task. Specifically, we present and analyze the core of scene graph guided MMT from the perspective of the encoder-decoder architecture in Section~\ref{sec:sgmmt}. Then, we detail the concrete encoder-decoder implementation of PSG in Section~\ref{sec:enc} and Section~\ref{sec:dec}. Finally, we introduce the overall loss function in Section~\ref{sec:loss}.

\subsection{Scene Graph Guided MMT Paradiam}\label{sec:sgmmt}
Neural Machine Translation (NMT) systems are generally based on encoder-decoder architecture. Give the source sentence $S=(s_1,s_2,\cdots,s_m)$ and the target sentence $T=(t_1,t_2,\cdots,t_n)$, the model $\mathcal{F}=(\mathcal{F}_{\rm enc},\mathcal{F}_{\rm dec})$ models the conditional likelihood of generating the target sequence as follows:
\begin{equation}
    \begin{aligned}
        P(T|S;\mathcal{F}) &= \prod_{i=1}^{n} \mathcal{F}(t_i|t_{<i}, S) \\
        &= \prod_{i=1}^{n} \mathcal{F}_{\rm dec}(t_i|t_{<i}, \mathcal{F}_{\rm enc}(S)),
    \end{aligned}
\end{equation}
where the decoder $\mathcal{F}_{\rm dec}$ takes the encoded representation of the source sentence $\mathcal{F}_{\rm enc}(S)$ along with the previously predicted target tokens $t_{<i}$ to generate the probability distribution over the target vocabulary.  The model is typically trained using the cross-entropy loss:
\begin{equation}
    \mathcal{L}_{\rm nmt} = \mathbb{E}_{(S,T)}[ - \log P(T| S;\mathcal{F})].
\end{equation}

Scene graph guided MMT extends this traditional NMT framework by incorporating multimodal information through visual and language scene graphs. Specifically, a visual scene graph $G_v$, generated from the image $I$, and a language scene graph $G_l$, derived from the source sentence $S$, are used to encode richer entitiy and relation information. This approach mitigates ambiguities that arise in text-only models by leveraging structured representations from both modalities.

The multimodal extension modifies the likelihood of generating the target sentence $T$ to:
\begin{equation}
    P(T|S;\mathcal{F}) = \prod_{i=1}^{n} \mathcal{F}(t_i|t_{<i}, \mathcal{F}_{\rm enc}(S, G_v, G_l)).
    \label{eq:sgmmt}
\end{equation}

Similarly, the loss function of MMT model can be formulated as:
\begin{equation}
    \mathcal{L}_{\rm mmt} = \mathbb{E}_{(S,T,I)}[ - \log P(T| S,I;\mathcal{F})].
\end{equation}

\subsection{Encoding Workflow}
\label{sec:enc}
The encoder $\mathcal{F}_{\rm enc}$ in our PSG framework includes four parts: text tokenization and embedding, scene graph extraction, scene graph pruning, and Transformer block joint encoding.
\subsubsection*{Text Tokenization and Embedding}
For the source sentence $S$, we first tokenize them using Byte Pair Encoding (BPE)~\cite{sennrich-etal-2016-neural} to produce tokenized sequences. These sequences are subsequently embedded into vector representations $\boldsymbol{f}_s \in \mathbb{R}^{m \times d_s}$ using an embedding layer.
\subsubsection*{Scene Graph Extraction}
To enhance the contextual understanding of source sentence $S$, we augment the textual representation with scene graph information. For the language modality information, we utilize the Stanford Language Scene Graph Parser (LSGP)~\cite{wang-etal-2018-scene} to capture relations between entities within $S$. The resulting language scene graph $G_l$ is defined as:
\begin{equation}
     \begin{aligned}
        G_l&={\rm LSGP}(S)\\
        &=\left\{E_l\in\mathbb{R}^{p_l\times1},R_l\in \mathbb{R}^{q_l\times1},{A}_l\in\mathbb{R}^{q_l\times2} \right\},
     \end{aligned}
\end{equation}
where $E_l$, $R_l$, and $A_l$ represent entity labels, relation labels, and the relation index matrix, respectively. Here, $p_l$ and $q_l$ denote the number of entities and relations in $G_l$.

For the visual information, we use the causal motifs Visual Scene Graph Network (VSGN)~\cite{tang2020unbiased} to extract the visual scene graph $G_v$ from the input image $I$:
\begin{equation}
     \begin{aligned}
        G_v&={\rm VSGN}(I)\\
        &=\left\{E_v\in\mathbb{R}^{p_v\times1},R_v\in \mathbb{R}^{q_v\times1},{A}_v\in\mathbb{R}^{q_v\times2} \right\},
     \end{aligned}
\end{equation}
where $E_v$, $R_v$, and $A_v$ represent the entity labels, relation labels, and relation index matrix of the visual scene graph. Similarly, $p_v$ and $q_v$ are the counts of entities and relations in $G_v$.

For the discrete entity and relation labels in scene graphs, we use CLIP model~\cite{radford2021learning} to embed these labels into continuous vectors. Notably, for the entitiy information in the visual scene graph $G_v$, instead of vectorize the entitiy labels using CLIP encoder, we straightly use the last hidden state in the object detection network (part of the VSGN) to represent the entities to retain more visual information. The vectorized scene graph $G_l^*$ and $G_v^*$ can be represented as:
\begin{equation}
    \begin{aligned}
        G_l^* &= \left\{E_{l}^*\in\mathbb{R}^{p_l\times d_c},R_{l}^*\in \mathbb{R}^{q_l\times d_c},{A}_l\in\mathbb{R}^{q_l\times2} \right\},\\
        G_v^* &= \left\{E_{v}^*\in\mathbb{R}^{p_v\times d_v},R_{v}^*\in \mathbb{R}^{q_v\times d_c},{A}_v\in\mathbb{R}^{q_v\times2} \right\}.
    \end{aligned}
\end{equation}

\subsubsection*{Scene Graph Message Passing and Pruning}
To aggregate information within the scene graphs, we employ the Multi-Layer Perceptrons (MLPs) to project both entity and relation vectors into a shared latent space of  dimension $d$. This is followed by the application of Graph Convolutional Networks (GCNs) to propagate information between entities and their relations. The update function for the $j$-th node ($i\in [1,p_l]$) in the language scene graph $G_l^*$ can be represented as:
\begin{equation}
    S(j, k) = \frac{
            \mathbf{W}_1 R_{l}^*[k] 
            + 
            \mathbf{W}_2 R_v^*[\tilde{A_l}(j, k)]
            }{\sqrt{\deg(k)\deg(j)}} 
        ,
\end{equation}
\begin{equation}
    \boldsymbol{f}_{l} = 
        \Bigg\{
        \sum_{k \in \mathcal{N}(j) \cup \{j\}} 
        \big[ S(j, k) + \mathbf{b} \big]
        \Bigg\}_{j=1}^{p_l}
        \in \mathbb{R}^{p_l \times d},
\end{equation}
where $\mathbf{W}_1$, $\mathbf{W}_2$, and $\mathbf{b}$ are all learnable parameters. The degree and neighbors of a node are represented as $\deg(\cdot)$ and $\mathcal{N}(\cdot)$, respectively. $\tilde{A_l}(\cdot)$ denotes the inverse index matrix of $A_l$. The above describes the message passing process for the language scene graph. Similarly, the aggregated representation $\boldsymbol{f}_v\in \mathbb{R}^{p_v \times d}$ for the visual scene graph can be computed using the same approach.

As pointed out in Section~\ref{sec:introduction}, visual scene graphs often contain an overabundance of nodes, many of which are irrelevant to the translation task. Such redundancy not only compromises translation accuracy but also imposes an additional computational burden on the backbone model. To address this, we propose leveraging the language scene graph to guide the pruning of the visual scene graph.

The cross-modal attention scores are first computed to measure the relevance between nodes of the visual and language scene graphs:
\begin{equation}
    \alpha_{v,l}[i,j]=\frac{\exp(\boldsymbol{f}_v[i]\cdot\boldsymbol{f}_l[j])}{\sum_{k=1}^{p_v}\exp(\boldsymbol{f}_v[i]\cdot\boldsymbol{f}_l[k])}.\\
\end{equation}

Next, the mean attention score for each node in the visual scene graph is obtained by aggregating its relevance across all nodes in the language scene graph:
\begin{equation}
\bar{\alpha}_v[i] = \frac{1}{p_l} \sum_{j=1}^{p_l} \alpha_{v,l}[i,j],
\end{equation}
where $\bar{\alpha}_v[i]$ represents the mean attention score of the $i$-th visual scene graph node. We then prune the visual scene graph by removing nodes with a mean attention score below a hyperparameter threshold $\tau$:
\begin{equation}
\boldsymbol{f}_v^{\prime} = \{i \in [1, p_v] | \bar{\alpha}_v[i] \geq  \frac{\tau}{p_v} \sum_{k=1}^{p_v} \bar{\alpha}_v[k]\}.
\end{equation}

Due to the irreversible nature of the pruning process, overly aggressive pruning could result in the loss of critical information. To mitigate this, we introduce a multi-step pruning strategy, where each step enforces a Kullback-Leibler divergence constraint between the visual scene graph and the language scene graph. Moreover, the pruning intensity is incrementally increased with each step to expedite convergence. The pruning loss function for step $\lambda$ can be expressed as:
\begin{equation}
\mathcal{L}_{\text{prune}} =\sum_{i=1}^{\lambda} \lambda\cdot{\rm KL}(\boldsymbol{f}_v^{(\lambda)}|| \boldsymbol{f}_l).
\end{equation}

\subsubsection*{Transformer Block Joint Encoding}
After Obtaining the plain text embedding $\boldsymbol{f}_s$, the language scene graph representation $\boldsymbol{f}_l$, and the pruned visual scene graph representation $\boldsymbol{f}_v^{(\lambda)}$, we integrate the multimodal information using an $L$-layer Transformer encoder, denoted as TRFM-E. The joint multimodal representation $\boldsymbol{f}_{\rm enc}$ is computed as follows:
\begin{equation}
\boldsymbol{f}_{\rm enc} = {\rm TRFM\text{-}E}([\boldsymbol{f}_s+{\rm PE}(\boldsymbol{f}_s); \boldsymbol{f}_l; \boldsymbol{f}_v^{(\lambda)}]).
\end{equation}
where ${\rm PE}(\cdot)$ denotes the positional encoding function, and $[\ ;\ ]$ denotes the concatenation operation.

\subsection{Decoding Workflow}
\label{sec:dec}
Consistent with Equation~\ref{eq:sgmmt}, we apply an $L$-layer Transformer decoder TRFM-D to perform autoregressive decoding on the joint representation $\boldsymbol{f}$ as follows:
\begin{equation}
\boldsymbol{f}_{\rm dec} = {\rm TRFM\text{-}D}([\boldsymbol{f}_{\rm enc}; \boldsymbol{f}_{\rm dec}^{(t-1)}])_{t=1}^{T}.
\end{equation}
The decoder output $\boldsymbol{f}{\rm dec}$ is then used to reconstruct the predicted sentence $T^*$ through a detokenization process. The multimodal machine translation loss $\mathcal{L}{\rm mmt}$ is computed by comparing $T^*$ with the ground truth.

\subsection{Overall Optimization}
\label{sec:loss}
The core losses of our PSG model include the MMT loss $\mathcal{L}_{\rm mmt}$ and the scene graph pruning loss $\mathcal{L}_{\rm prune}$. Moreover, due to the significant learning challenges imposed by multimodal data on the Transformer encoder-decoder modules, the model struggles to adapt to multimodal data from the very beginning. To address this issue, we propose adding an additional text-only NMT loss on top of the MMT loss, enabling the model to gradually adapt to multimodal training. The final loss function $\mathcal{L}$ is defined as:
\begin{equation}
\mathcal{L} = \mathcal{L}_{\rm mmt} +  \mathcal{L}_{\rm prune} + \mathcal{L}_{\rm nmt}.
\end{equation}

%% file: figures/overview.tex
\begin{figure*}[t]
    \centering
    \includegraphics[width=1\linewidth]{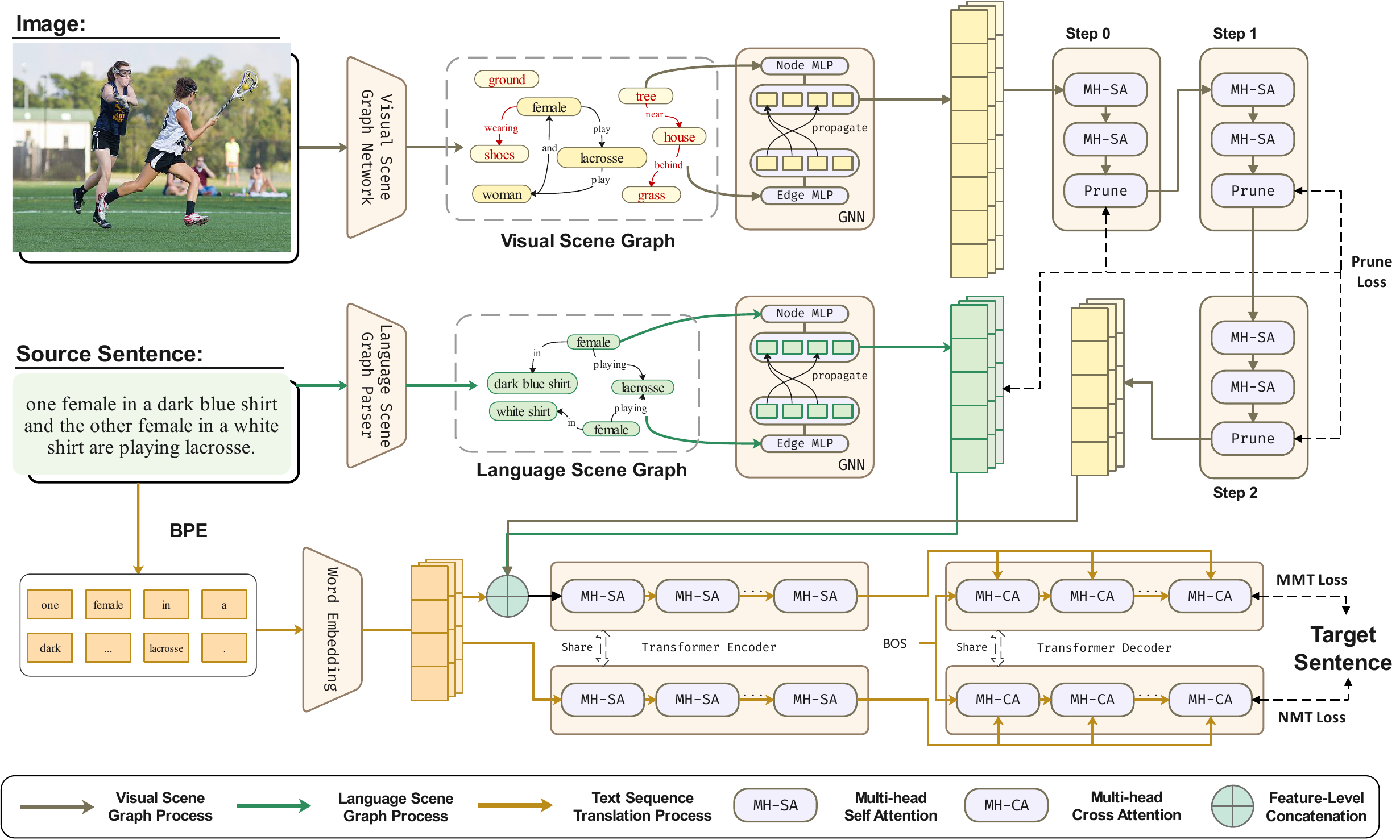}
    \caption{Overview of the PSG. The PSG framework consists of five key components: text sequence tokenization and embedding, scene graph extraction, scene graph pruning, joint representation encoding, and text decoding.}
    \label{fig:overview}
  \end{figure*}

%% file: sections/4_experiments.tex
\input{tables/sota_ende.tex}
\input{tables/sota_enfr.tex}
\section{Experiments}
\subsection{Experimental Settings}
\minisection{Datasets}
We evaluate the PSG model on three datasets: Multi30K~\cite{elliott-etal-2016-multi30k}, AmbigCaps~\cite{li-etal-2021-vision}, and CoMMuTE~\cite{futeral-etal-2023-tackling}, covering six tasks: En-De and En-Fr (Multi30K), En-Tr and Tr-En (AmbigCaps), and En-De and En-Fr (CoMMuTE). As CoMMuTE provides only a test set, we directly evaluate the model trained on Multi30K. Due to the prevalence of ambiguous texts in the AmbigCaps and CoMMuTE datasets, they are well-suited for evaluating a model’s ability to leverage visual information. (See Section~\ref{sec:appendix_dataset} for dataset sizes and details.)

\minisection{Evaluation Metrics}
We evaluate the PSG's performance using the BLEU-4 (hereafter referred to as BLEU)~\cite{papineni-etal-2002-bleu}, METEOR~\cite{denkowski-lavie-2014-meteor}, COMET~\cite{rei-etal-2020-comet}, and Accuracy~\cite{futeral-etal-2023-tackling}.

\minisection{Implementation Details}
The PSG model consists of a pretrained visual scene graph extraction network, a pretrained language scene graph extraction parser, a scene graph vectorization network, and a MMT backbone built using a Transformer architecture. For visual scene graph extraction, we utilize the Causal Motifs model~\cite{tang2020unbiased}, while language scene graph extraction is handled by the Stanford scene graph parser~\cite{wang-etal-2018-scene}. The scene graph vectorization process leverages CLIP~\cite{radford2021learning}, producing embeddings with a vector dimension of $512$. The Transformer backbone includes $6$ encoder and decoder layers, each with a hidden layer size of $512$, a feed-forward network intermediate size of $2048$, and $8$ attention heads.
\input{tables/sota_ab}
\input{tables/ablation.tex}

Our implementation is based on the Fairseq library~\cite{ott-etal-2019-fairseq}. For optimization, we use the Adam optimizer~\cite{kingma2014adam} with parameters $\beta_1= 0.9$, $\beta_2= 0.98$, and $\epsilon=10^{-8}$. The learning rate is set to $0.005$, with a $2000$-step warmup phase. The model employs a dropout rate of $0.3$ and a label smoothing coefficient of $0.1$. Notably, the pruning hyperparameters $\lambda$ and $\tau$ are set to $5$ and $0.2$, respectively, which yield the best experimental results.
\subsection{Comparisons with SOTA Methods}
In this section, we perform a comparative analysis between PSG and a branch of state-of-the-art MMT methods, including UVR-NMT~\cite{zhang2020neural}, Gated Fusion~\cite{wu-etal-2021-good}, MMT-VQA~\cite{zuo-etal-2023-incorporating}, and ConVisPiv~\cite{guo2024multi}, \textit{etc}. Additionally, we include results from a text-only machine translation model, Transformer~\cite{vaswani2017attention}, to highlight the significance of incorporating multimodal information. For fairness, methods using pretrained vocabularies and large parameters~\cite{gupta2023cliptrans} are excluded here. Their results are reported in Section~\ref{sec:pt}.

\subsubsection{Results on Multi30K}
Table~\ref{tab:sota_ende} presents the translation quality scores for the Multi30K English-German task. Our PSG model establishes new state-of-the art results, surpassing the previous best method, VALHALLA, by an average of $+1.84$ BLEU and $+0.84$ METEOR. It also surpasses all competing methods in terms of COMET scores. Furthermore, PSG demonstrates its superiority over the text-only Transformer model, achieving improvements of $+3.22$ in BLEU and $+1.76$ in METEOR. These results underscore the effectiveness of leveraging multimodal information to enhance translation quality.

Similarly, the PSG also excels in English-French translation, as shown in Table~\ref{tab:sota_enfr}. It attains the highest average BLEU score of $56.65$ and METEOR score of $77.31$, outperforming VALHALLA by $+1.48$ in BLEU and $+0.58$ in METEOR.

\subsubsection{Results on AmbigCaps and CoMMuTE}
To further verify the model’s robustness in complex semantic scenarios, we conduct evaluations on two challenging ambiguity-focused datasets: AmbigCaps and CoMMuTE. As shown in Table~\ref{tab:sota_ab}, the PSG model achieves the best performance across nearly all evaluation metrics. These results demonstrate that the proposed visual scene graph pruning strategy effectively leverages visual information to resolve linguistic ambiguities, showing significant advantages in MMT tasks.

\input{tables/param_p.tex}
\input{figures/attention.tex}
\subsection{Ablation Studies}
To validate the effectiveness of the scene graph pruning module and text-only NMT constraint in the PSG model, we conduct ablation studies on the Multi30K. As listed in Table~\ref{tab:ablation}, we observe that the text-only NMT constraint significantly improves the translation quality, with an increase of $+1.16$ BLEU, $+1.12$ METEOR, $+0.033$ COMET on En-De, and $+1.07$ BLEU, $+0.72$ METEOR, and $+0.022$ on En-Fr compared to the baseline.

In contrast, the improvement achieved by using the scene graph pruning module alone is not very significant. However, when combined with the text-only NMT constraint, it produces a synergistic effect, achieving best performance. These results are consistent with our discussion in Section~\ref{sec:loss}, where we argue that incorporating multimodal information increases the burden on the original Transformer encoder-decoder backbone, making the addition of text-only constraints essential.

Furthermore, we introduce a random pruning strategy as a baseline. However, due to the absence of linguistic guidance, its performance not only falls short of our proposed pruning method but is even worse than applying no pruning at all.

\subsection{Sensitivity Studies}
In this section, we conduct sensitivity studies to assess the impact of two key coefficients, \textit{i.e.}, the pruning steps $\lambda$ and the pruning threshold $\tau$. The performance variations corresponding to different values of $\lambda$ and $\tau$ are presented in Table~\ref{tab:sensitivity}. Overall, both for the En-De and En-Fr translation tasks, the performance is optimal when $\lambda$ and $\tau$ are set to $5$ and $0.2$, respectively. Setting $\lambda$ and $\tau$ either too high or too low results in insufficient pruning strength or excessive information loss, leading to a decline in performance.

\subsection{Visualizations}
To further validate the effectiveness of our scene graph pruning module, we provide visualizations of attention score maps for visual and language scene graph entities, both before and after pruning, in Figure~\ref{fig:attention}. As shown, guided by the language scene graph, PSG successfully removes redundant entity nodes from the visual scene graph, such as ``hat'' in Figure~\ref{fig:attention}(a) and ``short'' in Figure~\ref{fig:attention}(b), preventing noise from adversely impacting downstream translation performance.

%% file: tables/sota_ende.tex
\begin{table*}[ht]
    \centering
    \setlength\tabcolsep{4pt}
    \resizebox{1\linewidth}{!}{
    \begin{tabular}{l|ccc|ccc|ccc|ccc}
    \toprule
    \multirow{2}{*}{\textbf{Methods}} & \multicolumn{3}{c|}{\textbf{Test 2016}} & \multicolumn{3}{c|}{\textbf{Test 2017}} & \multicolumn{3}{c|}{\textbf{MSCOCO}} & \multicolumn{3}{c}{\textbf{Average}} \\
    \cmidrule(lr){2-13}
     & BLEU & METEOR &COMET& BLEU & METEOR &COMET& BLEU & METEOR &COMET& BLEU & METEOR &COMET \\
    \midrule
    \midrule
    \multicolumn{9}{l}{\textbf{Text-only}} \\
    \midrule
    Transformer~\citeyearpar{vaswani2017attention} & $41.02$ & $68.22$ &$-$& $33.36$ & $62.05$&$-$ & $29.88$ & $56.65$ &$-$& $34.75$ & $62.31$&$-$ \\
    \midrule
    \multicolumn{9}{l}{\textbf{Multimodal}} \\
    \midrule
    UVR-NMT~\citeyearpar{zhang2020neural} & $40.79$ & $-$ &$-$& $32.16$ & $-$ &$-$& $29.02$ & $-$ &$-$& $33.99$ & $-$ &$-$\\
    Graph-MMT~\citeyearpar{yin-etal-2020-novel} & $39.80$ & $57.60$ &\textcolor{reproduce1}{\underline{$0.368$}}& $32.20$ & $51.90$ &\textcolor{reproduce1}{$0.226$}& $28.70$ & $47.60$ &\textcolor{reproduce1}{\underline{$0.060$}}& $33.57$ & $52.37$ &\textcolor{reproduce1}{$0.218$}\\
    Gated Fusion~\citeyearpar{wu-etal-2021-good} & $41.96$ & $67.84$ &\textcolor{reproduce1}{$\mathbf{0.378}$}& $33.59$ & $61.94$ &\textcolor{reproduce1}{\underline{$0.236$}}& $29.04$ & $56.15$ &\textcolor{reproduce1}{$0.055$}& $34.86$ & $61.98$ &\textcolor{reproduce1}{\underline{$0.223$}}\\
    VALHALLA~\citeyearpar{li2022valhalla} & $42.60$ & $\mathbf{69.30}$ &$-$& $35.10$ & \underline{$62.80$} &$-$& $30.70$ & \underline{$57.60$} &$-$& $36.13$ & \underline{$63.23$} &$-$\\
    PLUVR~\citeyearpar{fang-feng-2022-neural} & $40.30$ & $-$ &$-$& $33.45$ & $-$ &$-$& $30.28$ & $-$ &$-$& $34.68$ & $-$ &$-$\\
    IVA~\citeyearpar{ji-etal-2022-increasing} & $41.77$ & $68.60$ &$-$& $34.58$ & $62.40$ &$-$& $30.61$ & $56.70$ &$-$& $35.65$ & $62.57$ &$-$\\
    Selective Attn~\citeyearpar{li-etal-2022-vision} & $41.93$ & $68.55$ &$-$& $33.60$ & $61.42$ &$-$& $31.14$ & $56.77$ &$-$& $35.56$ & $62.25$ &$-$\\
    IKD-MMT~\citeyearpar{peng-etal-2022-distill} & $41.28$ & $-$ &$-$& $33.83$ & $-$ &$-$& $30.17$ & $-$ &$-$& $35.09$ & $-$ &$-$\\
    MMT-VQA~\citeyearpar{zuo-etal-2023-incorporating} & $42.55$ & $69.00$ &$-$& $34.58$ & $61.99$ &$-$& $30.96$ & $57.23$ &$-$& $36.03$ & $62.74$ &$-$\\
    SAMMT~\citeyearpar{guo-etal-2023-bridging} & $42.50$ & $-$ &$-$& \underline{$36.04$} & $-$ &$-$& \underline{$31.95$} & $-$ &$-$& \underline{$36.83$} & $-$ &$-$\\
    RG-MMT-EDC~\citeyearpar{tayir2024encoder} & $42.00$ & $60.20$ &$-$& $33.40$ & $53.70$ &$-$& $30.00$ & $49.60$ &$-$& $35.13$ & $54.50$ &$-$\\
    ConVisPiv~\citeyearpar{guo2024multi} & \underline{$42.64$} & $60.56$ &$-$& $34.84$ & $54.62$ &$-$& $29.69$ & $50.12$ &$-$& $35.72$ & $55.10$ &$-$\\
    \midrule
    \rowcolor{black!6}
    PSG & $\mathbf{42.75}$ & \underline{$69.11$} &$0.351$& $\mathbf{37.50}$ & $\mathbf{63.87}$ &$\mathbf{0.256}$& $\mathbf{33.66}$ & $\mathbf{59.22}$ &$\mathbf{0.169}$& $\mathbf{37.97}$ & $\mathbf{64.07}$ &$\mathbf{0.259}$\\
    \bottomrule
    \end{tabular}}
    \caption{Comparisons with SOTA methods on the Multi30K English-German benchmark. The \textcolor{reproduce1}{blue} results are reproduced by~\citet{futeral-etal-2023-tackling}. The numbers in \textbf{bold} represent the top-performing results, while the \underline{underlined} numbers indicate the second-best outcomes.}
    \label{tab:sota_ende}
    \end{table*}

%% file: tables/sota_enfr.tex
\begin{table*}[ht]
    \centering
    \setlength\tabcolsep{4pt}
    \resizebox{1\linewidth}{!}{
    \begin{tabular}{l|ccc|ccc|ccc|ccc}
    \toprule
    \multirow{2}{*}{\textbf{Methods}} & \multicolumn{3}{c|}{\textbf{Test 2016}} & \multicolumn{3}{c|}{\textbf{Test 2017}} & \multicolumn{3}{c|}{\textbf{MSCOCO}} & \multicolumn{3}{c}{\textbf{Average}} \\
    \cmidrule(lr){2-13}
     & BLEU & METEOR &COMET & BLEU & METEOR&COMET & BLEU & METEOR &COMET& BLEU & METEOR&COMET \\
    \midrule
    \midrule
    \multicolumn{13}{l}{\textbf{Text-only}} \\
    \midrule
    Transformer~\citeyearpar{vaswani2017attention} & $61.80$ & $81.02$ &$-$& $53.46$ & $75.62$&$-$ & $44.52$ & $69.43$&$-$ & $53.26$ & $75.36$&$-$ \\
    \midrule
    \multicolumn{13}{l}{\textbf{Multimodal}} \\
    \midrule
    Imagination~\citeyearpar{elliott-kadar-2017-imagination} & $61.90$ & $-$&$-$ & $54.85$ & $-$&$-$ & $44.86$ & $-$&$-$ & $53.80$ & $-$&$-$ \\

    Graph-MMT~\citeyear{yin-etal-2020-novel}&$60.90$&$74.90$&\textcolor{reproduce1}{$0.705$}&$53.90$&$69.30$&\textcolor{reproduce1}{\underline{$0.589$}}&$-$&$-$&\textcolor{reproduce1}{$0.387$}&$-$&$-$&\textcolor{reproduce1}{$0.560$}\\
    Gated Fusion~\citeyearpar{wu-etal-2021-good} & $61.69$ & $80.97$&\textcolor{reproduce1}{\underline{$0.707$}} & $54.85$ & $76.34$&\textcolor{reproduce1}{$0.580$} & $44.86$ & $70.51$&\textcolor{reproduce1}{\underline{$0.394$}} & $53.80$ & $75.94$&\textcolor{reproduce1}{\underline{$0.560$}} \\

    VALHALLA~\citeyearpar{li2022valhalla} & \underline{$63.10$} & \underline{$81.80$} &$-$& \underline{$56.00$} & \underline{$77.10$} &$-$& $46.40$ & \underline{$71.30$} &$-$& \underline{$55.17$} & \underline{$76.73$} &$-$\\

    PLUVR~\citeyearpar{fang-feng-2022-neural} & $61.31$ & $-$ &$-$& $53.15$ & $-$ &$-$& $43.65$ & $-$ &$-$& $52.70$ & $-$ & $-$\\

    Selective Attn~\citeyearpar{li-etal-2022-vision} & $62.48$ & $81.71$ &$-$& $54.44$ & $76.46$ &$-$& $44.72$ & $71.20$ &$-$& $53.88$ & $76.46$ \\

    IKD-MMT~\citeyearpar{peng-etal-2022-distill} & $62.53$ & $-$ &$-$& $54.84$ & $-$&$-$ & $-$ & $-$&$-$ & $-$ & $-$ & $-$\\

    MMT-VQA~\citeyearpar{zuo-etal-2023-incorporating} & $62.24$ & $81.77$ &$-$& $54.89$ & $76.53$ &$-$& $45.75$ & $71.21$ &$-$& $54.29$ & $76.50$&$-$ \\

    RG-MMT-EDC~\citeyearpar{tayir2024encoder} & $62.90$ & $77.20$ &$-$& $55.80$ & $72.00$ &$-$& $45.10$ & $64.90$ &$-$& $54.60$ & $71.37$ &$-$\\

    ConVisPiv~\citeyearpar{guo2024multi} & $62.56$ & $77.09$ &$-$& $55.83$ & $73.18$ &$-$& \underline{$46.61$} & $67.67$ &$-$& $55.00$ & $72.65$&$-$ \\

    \midrule
    \rowcolor{black!6}
    PSG & $\mathbf{64.22}$ & $\mathbf{82.27}$ &$\mathbf{0.739}$& $\mathbf{57.66}$ & $\mathbf{77.73}$ &$\mathbf{0.698}$& $\mathbf{48.06}$ & $\mathbf{71.93}$&$\mathbf{0.549}$ & $\mathbf{56.65}$ & $\mathbf{77.31}$ &$\mathbf{0.662}$\\
    \bottomrule
    \end{tabular}}
    \caption{Comparisons with SOTA methods on the Multi30K English-French benchmark.}
    \label{tab:sota_enfr}
    \end{table*}

%% file: tables/sota_ab.tex
\begin{table*}[ht]
    \centering
    \setlength\tabcolsep{8pt}
    \resizebox{1\linewidth}{!}{
    \begin{tabular}{l|ccc|ccc|c|c}
    \toprule
    \multirow{2}{*}{\textbf{Methods}} & \multicolumn{3}{c|}{\textbf{AmbigCaps En$\rightarrow$Tr}} & \multicolumn{3}{c|}{\textbf{AmbigCaps Tr$\rightarrow$En}} & \multicolumn{1}{c|}{\textbf{CoMMuTE En$\rightarrow$De}} & \multicolumn{1}{c}{\textbf{CoMMuTE En$\rightarrow$Fr}} \\
    \cmidrule(lr){2-9}
     & BLEU & METEOR &COMET& BLEU & METEOR &COMET& Accuracy & Accuracy\\
    \midrule
    \midrule
    \multicolumn{9}{l}{\textbf{Text-only}} \\
    \midrule

    Transformer~\citeyearpar{vaswani2017attention} & \textcolor{reproduce3}{$28.84$} & \textcolor{reproduce3}{$55.06$} &\textcolor{reproduce3}{$0.464$}& \textcolor{reproduce3}{$36.29$} & \textcolor{reproduce3}{$66.97$}&\textcolor{reproduce3}{$0.339$} & \textcolor{reproduce1}{$50.0$} & \textcolor{reproduce1}{$50.0$} \\
    
    \midrule
    \multicolumn{9}{l}{\textbf{Multimodal}} \\
    \midrule

    Graph-MMT~\citeyearpar{yin-etal-2020-novel} & $-$ & $-$ &$-$& $-$ & $-$ &$-$& \textcolor{reproduce1}{$49.1$} & \textcolor{reproduce1}{$50.2$} \\
    Gated Fusion~\citeyearpar{wu-etal-2021-good} & \textcolor{reproduce3}{$36.47$} & \textcolor{reproduce3}{$61.29$} &\textcolor{reproduce3}{$0.641$}& \textcolor{reproduce3}{$41.81$} & \textcolor{reproduce3}{$70.74$} &\textcolor{reproduce3}{$0.428$}& \textcolor{reproduce1}{$49.7$} & \textcolor{reproduce1}{$50.0$}\\
    Concatenation~\citeyearpar{li-etal-2021-vision} & $-$ & $-$ &$-$& $37.39$ & $-$ &$-$& $-$ & $-$\\
    \midrule
    \rowcolor{black!6}
    PSG & $\mathbf{36.86}$ & $\mathbf{62.42}$ &$\mathbf{0.692}$& $\mathbf{42.09}$ & $\mathbf{71.15}$ &$\mathbf{0.447}$& $\mathbf{51.0}$ & $\mathbf{50.6}$ \\
    \bottomrule
    \end{tabular}}
    \caption{Comparisons with SOTA methods on AmbigCaps and CoMMuTE. The \textcolor{reproduce3}{red} results are reproduced by us.}
    \label{tab:sota_ab}
    \end{table*}

%% file: tables/ablation.tex
\begin{table*}[t]
    \centering
    \setlength\tabcolsep{1pt}
    \resizebox{1\linewidth}{!}{
    \begin{tabular}{c|cc|ccc|ccc|ccc|ccc|ccc}
    \toprule
    \multirow{2}{*}{\textbf{Method}} &\multirow{2}{*}{$\boldsymbol{\mathcal{L}_{\rm prune}}$} &\multirow{2}{*}{$\boldsymbol{\mathcal{L}_{\rm nmt}}$} & \multicolumn{3}{c|}{\textbf{Test 2016}} & \multicolumn{3}{c|}{\textbf{Test 2017}} &
    \multicolumn{3}{c|}{\textbf{Test 2018}} &
    \multicolumn{3}{c|}{\textbf{MSCOCO}} & \multicolumn{3}{c}{\textbf{Average}} \\
    \cmidrule(lr){4-18}
     &&& BLEU & METEOR &COMET& BLEU & METEOR&COMET & BLEU & METEOR&COMET &BLEU & METEOR&COMET &BLEU & METEOR&COMET \\
    \midrule
    \midrule
    \multicolumn{18}{l}{En$\rightarrow$De}\\
    \midrule
    \multirow{5}{*}{PSG} &\ding{55}&\ding{55}& $41.27$ & $67.73$ &$0.309$ & $34.89$ & $61.75$&$0.216$ &$33.59$ &$58.57$&$0.152$ & $30.98$ & $56.35$&$0.124$ & $35.18$ & $61.10$&$0.200$ \\

    &\ding{51}&\ding{55}& $41.53$ & $68.14$&$0.323$ & $34.66$ & $61.44$ &$0.211$&$32.94$ &$58.09$&$0.142$ & $31.72$ & $56.80$&$0.116$ & $35.21$ & $61.18$&$0.198$ \\
    
    &\ding{55}&\ding{51}& $\mathbf{42.81}$ & $69.00$&$0.347$ & $36.11$ & $62.93$&$0.248$ &$33.85$ &$59.25$&$0.183$ & $32.57$ & $57.70$&$0.155$ & $36.34$ & $62.22$&$0.233$ \\

    &\scalebox{0.85}{\Checkmark\kern-1.2ex\raisebox{1ex}{\rotatebox[origin=c]{125}{\textbf{--}}}}&\ding{51}& $42.38$ & $68.53$ &$0.344$& $35.85$ & $62.51$ &$0.232$&$33.63$ &$59.05$&$0.176$ & $32.28$ & $57.44$&$0.158$ & $36.04$ & $61.88$&$0.228$ \\

    &\cellcolor{black!6}\ding{51}&\cellcolor{black!6}\ding{51}& \cellcolor{black!6}$42.75$ & \cellcolor{black!6}$\mathbf{69.11}$ &\cellcolor{black!6}$\mathbf{0.351}$& \cellcolor{black!6}$\mathbf{37.50}$ & \cellcolor{black!6}$\mathbf{63.87}$ &\cellcolor{black!6}$\mathbf{0.256}$& \cellcolor{black!6}$\mathbf{34.55}$ & \cellcolor{black!6}$\mathbf{59.84}$ &\cellcolor{black!6}$\mathbf{0.196}$& \cellcolor{black!6}$\mathbf{33.66}$ & \cellcolor{black!6}$\mathbf{59.22}$ &\cellcolor{black!6}$\mathbf{0.169}$& \cellcolor{black!6}$\mathbf{37.11}$ & \cellcolor{black!6}$\mathbf{63.01}$ &\cellcolor{black!6}$\mathbf{0.243}$\\
    
    \midrule
    \multicolumn{18}{l}{En$\rightarrow$Fr}\\
    \midrule
    
    \multirow{5}{*}{PSG} &\ding{55}&\ding{55}& $63.19$ & $81.52$ &$0.707$& $55.82$ & $76.62$&$0.658$ &$40.00$ &$64.24$&$0.541$ & $46.36$ & $70.93$&$0.523$ & $51.34$ & $73.33$&$0.607$ \\

    &\ding{51}&\ding{55}& $62.99$ & $81.53$&$0.717$ & $55.87$ & $76.53$&$0.654$ &$40.21$ &$64.80$ &$0.548$& $46.50$ & $70.63$&$0.527$ & $51.39$ & $73.37$&$0.612$ \\

    &\ding{55}&\ding{51}& $63.77$ & $82.10$&$0.731$ & $57.02$ & $77.41$&$0.683$ &$\mathbf{41.47}$ &$65.00$&$0.557$ & $47.76$ & $71.71$ &$0.544$& $52.41$ & $74.05$&$0.629$ \\

    &\scalebox{0.85}{\Checkmark\kern-1.2ex\raisebox{1ex}{\rotatebox[origin=c]{125}{\textbf{--}}}}&\ding{51}& $63.21$ & $81.74$ &$0.728$& $56.63$ & $77.07$ &$0.689$&$40.98$ &$64.56$ &$0.543$& $47.19$ & $71.37$ &$0.531$& $52.00$ & $73.69$&$0.623$ \\

    &\cellcolor{black!6}\ding{51}&\cellcolor{black!6}\ding{51}& \cellcolor{black!6}$\mathbf{64.22}$ & \cellcolor{black!6}$\mathbf{82.27}$ &\cellcolor{black!6}$\mathbf{0.739}$& \cellcolor{black!6}$\mathbf{57.66}$ & \cellcolor{black!6}$\mathbf{77.73}$ &\cellcolor{black!6}$\mathbf{0.689}$&\cellcolor{black!6}$41.25$ & \cellcolor{black!6}$\mathbf{65.06}$ &\cellcolor{black!6}$\mathbf{0.559}$& \cellcolor{black!6}$\mathbf{48.06}$ & \cellcolor{black!6}$\mathbf{71.93}$ &\cellcolor{black!6}$\mathbf{0.549}$& \cellcolor{black!6}$\mathbf{52.80}$ & \cellcolor{black!6}$\mathbf{74.24}$ &\cellcolor{black!6}$\mathbf{0.636}$\\

    \bottomrule
    \end{tabular}}
    \caption{The ablation results concerning the scene graph pruning module and text-only neural machine translation constraint of PSG on the Multi30K English-German and English-French benchmarks. $\mathcal{L}_{\rm prune}$ being \scalebox{0.85}{\Checkmark\kern-1.2ex\raisebox{1ex}{\rotatebox[origin=c]{125}{\textbf{–}}}} indicates that the pruning of visual information is performed using a random selection strategy.}
    \label{tab:ablation}
    \end{table*}

%% file: tables/param_p.tex
\begin{table*}[t]
    \centering
    \setlength\tabcolsep{6pt}
    \resizebox{1\linewidth}{!}{
    \begin{tabular}{c|cc|cc|cc|cc|cc|cc}
    \toprule
    \multirow{2}{*}{\textbf{Method}} &\multirow{2}{*}{$\boldsymbol{\lambda}$} &\multirow{2}{*}{$\boldsymbol{\tau}$} & \multicolumn{2}{c|}{\textbf{Test 2016}} & \multicolumn{2}{c|}{\textbf{Test 2017}} &
    \multicolumn{2}{c|}{\textbf{Test 2018}} &
    \multicolumn{2}{c|}{\textbf{MSCOCO}} & \multicolumn{2}{c}{\textbf{Average}} \\
    \cmidrule(lr){4-13}
     &&& BLEU & METEOR & BLEU & METEOR & BLEU & METEOR &BLEU & METEOR &BLEU & METEOR \\
    \midrule
    \midrule
    \multicolumn{13}{l}{En$\rightarrow$De}\\
    \midrule
    \multirow{6}{*}{PSG} &$0$& $-$&$42.81$ & $69.00$ & $36.11$ & $62.93$ &$33.85$ &$59.25$ & $32.57$ & $57.70$ & $36.34$ & $62.22$ \\

    &$3$&$0.2$& $43.22$ & $68.99$ & $35.90$ & $62.88$ &$34.56$ &$59.57$ & $33.13$ & $57.60$ & $36.70$ & $62.26$ \\
    
    &$5$&$0.1$&$42.30$ & $68.30$ & $36.29$ & $62.93$ &$34.36$ &$59.60$ & $33.01$ & $58.34$ & $36.49$ & $62.29$ \\

    & \cellcolor{black!6}$5$& \cellcolor{black!6}$0.2$& \cellcolor{black!6}$42.75$ & \cellcolor{black!6}$\mathbf{69.11}$ & \cellcolor{black!6}$\mathbf{37.50}$ & \cellcolor{black!6}$\mathbf{63.87}$ & \cellcolor{black!6}$34.55$ & \cellcolor{black!6}$59.84$ & \cellcolor{black!6}$\mathbf{33.66}$ & \cellcolor{black!6}$\mathbf{59.22}$ & \cellcolor{black!6}$\mathbf{37.11}$ & \cellcolor{black!6}$\mathbf{63.21}$ \\

    &$5$&$0.3$&$\mathbf{42.76}$ & $68.74$ & $36.68$ & $63.41$ &$\mathbf{35.67}$ &$\mathbf{60.34}$ & $33.06$ & $57.80$ & $37.04$ & $62.57$ \\

    &$7$&$0.2$&$42.59$ & $68.46$ & $36.24$ & $62.75$ &$35.21$ &$59.88$ & $33.06$ & $58.23$ & $36.77$ & $62.33$ \\
    \midrule
    \multicolumn{13}{l}{En$\rightarrow$Fr}\\
    \midrule

    \multirow{6}{*}{PSG}& $0$ &$-$ & $63.77$ & $82.10$ & $57.02$ & $77.41$ &$\mathbf{41.47}$ &$65.00$ & $47.76$ & $71.71$ & $52.41$ & $74.05$ \\

    &$3$&$0.2$ & $63.93$ & $82.24$ & $56.83$ & $77.20$ &$40.20$ &$64.35$ & $46.82$ & $71.55$ & $51.94$ & $73.83$ \\
    
    &$5$&$0.1$ & $64.20$ & $\mathbf{82.48}$ & $57.12$ & $77.68$ &$40.67$ &$64.71$ & $45.99$ & $70.98$ & $51.99$ & $73.96$ \\

    &\cellcolor{black!6}$5$&\cellcolor{black!6}$0.2$& \cellcolor{black!6}$\mathbf{64.22}$ & \cellcolor{black!6}$82.27$ & \cellcolor{black!6}$\mathbf{57.66}$ & \cellcolor{black!6}$\mathbf{77.73}$ &\cellcolor{black!6}$41.25$ &\cellcolor{black!6}$\mathbf{65.06}$ & \cellcolor{black!6}$\mathbf{48.06}$ & \cellcolor{black!6}$\mathbf{71.93}$ & \cellcolor{black!6}$\mathbf{52.80}$ & \cellcolor{black!6}$\mathbf{74.24}$ \\

    &$5$&$0.3$ & $63.87$ & $82.29$ & $57.07$ & $77.20$ &$40.89$ &$64.92$ & $47.04$ & $71.45$ & $52.21$ & $73.96$ \\

    &$7$&$0.2$ & $63.63$ & $82.17$ & $56.92$ & $77.42$ &$41.30$ &$65.06$ & $47.84$ & $71.56$ & $52.42$ & $74.05$ \\
    \bottomrule
    \end{tabular}}
    \caption{Sensitivity analysis results concerning the pruning steps $\lambda$ and the pruning threshold $\tau$ of PSG on the Multi30K English-German and English-French benchmarks.}
    \label{tab:sensitivity}
    \end{table*}

%% file: figures/attention.tex
\begin{figure*}[h!]
    \includegraphics[width=1\linewidth]{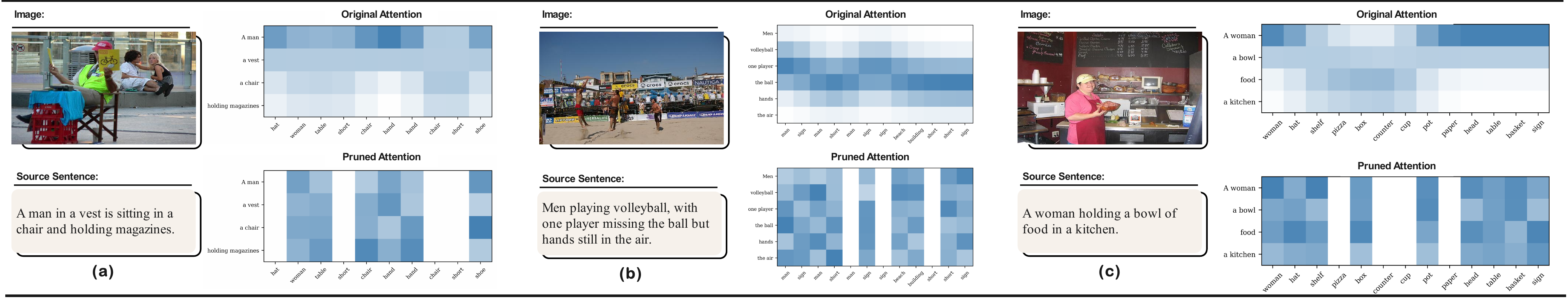}
    \caption{Comparison of attention visualization before and after pruning.}
    \label{fig:attention}
\end{figure*}

%% file: sections/5_conclusion.tex
\section{Conclusion}
MMT aims to address the challenges posed by linguistic polysemy and ambiguity in translation tasks by integrating image information. The current bottleneck in MMT research lies in the effective utilization of visual information. Previous approaches have extracted global or region-level image features and employed attention mechanisms or gating mechanisms for multimodal information fusion. However, these methods have failed to address the issue of visual information redundancy in MMT and propose effective solutions. In this paper, we introduce multimodal machine translation based on Scene Graph Pruning (PSG), which leverages language scene graph information to guide the pruning of redundant nodes in visual scene graphs, thereby reducing noise in downstream translation tasks. Moreover, we highlight the issue that multimodal data imposes significant learning pressure on the Transformer backbone, leading to low learning efficiency. To address this, we propose applying an additional text-only machine translation loss to guide multimodal learning. Extensive comparative experiments with state-of-the-art methods, along with comprehensive ablation studies, demonstrate the superior performance of the PSG model. These findings not only validate the effectiveness of our approach but also underscore the potential of visual information pruning as a promising direction for advancing multimodal machine translation research.

%% file: sections/6_limitation.tex
\section*{Limitations}
First, we rely on pre-trained scene graph extraction networks or parsers to generate scene graphs, which facilitate the translation of text sequences and improve the performance of the translation task. However, this dependency makes our method sensitive to the quality of these external networks. If the generated visual and language scene graphs contain significant noise, it may negatively affect our approach. Second, we have evaluated the adaptability of the scene graph pruning network using Transformers of varying sizes, showing that deeper and wider Transformer backbones yield better performance. However, this conclusion has not yet been validated on larger (billion-parameter) translation backbones, leaving this as a promising direction for future research.

%% file: sections/7_appendix.tex
\section{More Implementation Details}
\subsection{Dataset Details}\label{sec:appendix_dataset}
Multi30K is an extended version of the Flickr dataset and is currently one of the most widely used multimodal machine translation datasets. Each sample is associated with an image and corresponding English, German, and French descriptions. The training and validation sets contain $29,000$ and $1,014$ samples, respectively. To ensure experimental comparability, we follow previous studies and conduct evaluations on the Test2016, Test2017, Test2018, and MSCOCO test sets, which contain $1,000$, $1,000$, $1,071$, and $461$ instances, respectively.

The AmbigCaps and CoMMuTE datasets are specifically designed for multimodal machine translation tasks, aiming to evaluate a model’s ability to leverage visual information through a large number of ambiguous texts. The AmbigCaps dataset includes $89,601$ images for training, $1,000$ for validation, and $1,000$ for testing, with each image accompanied by corresponding English and Turkish descriptions. In contrast, the CoMMuTE dataset only contains a test set, with English-German and English-French parallel corpora consisting of $300$ and $308$ samples, respectively. Each sample provides a pair of target references (correct and incorrect), and the model is required to effectively utilize visual information to predict the correct answer.
\subsection{Training Procedure}
\input{tables/architecture.tex}
\input{tables/cost}
The PSG model is optimized using the Adam optimizer with an inverse square root learning rate schedule and warm-up steps. To ensure a fair comparison with prior studies, we adopt an early stopping mechanism, terminating training if validation performance does not improve within $10$ epochs. Key optimization hyperparameters are summarized in Table~\ref{tab:params} for reference.

Moreover, in Table~\ref{tab:cost}, we present the computational costs of all modules, including scene graph extraction, scene graph vectorization, and translation, to facilitate the assessment of the reproducibility of this work. Overall, the time cost for data preprocessing is mainly concentrated in the visual scene graph generation module. The additional cost of visual pruning during translation, measured by the difference between PSG and PSG w/o pruning, is $3.5\%$, resulting in a $+0.77$ BLEU improvement—a worthwhile trade-off.
\subsection{Inference and Evaluation}
For inference, we average the last $10$ checkpoints to achieve robust performance. We use beam search with a beam size of $5$
to generate translation outputs. The calulation of BLEU scores is based on Fairseq library and the calulation of METEOR scores is based on NLTK library~\cite{bird2009natural}.

The BLEU can be calulated as follows:
\begin{equation}
    {\rm BLEU} = (1-r)\times \exp\left(\frac{1}{N}\sum_{n=1}^N \log P_n\right)
\end{equation}
where $P_n$ is the n-gram precision, and $r$ is the ratio of reference length to prediction length.

The METEOR score is computed as follows:
\begin{equation}
    {\rm METEOR} = (1-\gamma\cdot F^\beta)\cdot\frac{R\cdot P}{\alpha P + (1-\alpha)R},
\end{equation}
where $R$ is the recall, $P$ is the precision, $F$ is the fragmentation fraction. $\alpha$, $\beta$, and $\gamma$ are hyperparameters that control the weights of precision and recall, the shape of the penalty, and the weight of the penalty term.

The COMET score is a neural network-based evaluation metric built upon pre-trained language models, capable of capturing richer semantic information. It generates scores that better align with human judgments by jointly considering the contextual representations of the source sentence, reference translation, and predicted translation. In our experiments, we adopt the evaluation model wmt20-comet-da.

The Accuracy metric, proposed by~\cite{futeral-etal-2023-tackling}, evaluates how well a model distinguishes between correct and incorrect translations based on perplexity. The core idea is to compare the relative closeness of the predicted translation to the correct and incorrect references. The computation is defined as:
\begin{equation}
{\rm Accuracy} = \mathbb{I} \left[ {\rm PPL}(T^*, T^+) < {\rm PPL}(T^*, T^-) \right],
\end{equation}
where $T^+$ and $T^-$ represent the correct and incorrect reference translations, respectively. ${\rm PPL}(\cdot,\cdot)$ is the perplexity function, and $\mathbb{I}[\cdot,\cdot]$ is the indicator function that returns $1$ if the condition holds, and $0$ otherwise.

\section{More Experimental Results}
\input{figures/bs_ende.tex}
\input{figures/bs_enfr.tex}
\input{tables/size.tex}

\subsection{Batch Size}
Figure~\ref{fig:bs_ende} and Figure~\ref{fig:bs_enfr} present the performance results of the PSG model trained with various batch sizes. The data reveals that both excessively large and excessively small batch sizes negatively impact the model's training effectiveness, espeically the small batch size. Notably, the model attains its best performance when the batch size is set to $4096$, indicating that this particular size strikes an optimal balance between stability and convergence speed.
\subsection{Transformer Backbone Size}
\input{tables/pt}
\input{figures/case.tex}
Table~\ref{tab:size} presents the performance results of the PSG model trained with different backbone sizes. The results indicate that the model's performance improves as the backbone size increases, with the largest backbone size achieving the best results. This suggests that larger backbone sizes can lead to better translation quality. Additionally, the learning rate should be appropriately reduced when using larger models; otherwise, the model’s performance may degrade.
\subsection{Pretrained Backbone Baselines}\label{sec:pt}

To better contextualize recent advances in MMT, we present a comparative analysis of several pretrained baselines in Table~\ref{tab:pt}. While these methods demonstrate superior performance by leveraging mBART's extensive vocabulary and robust sentence understanding capabilities, their effectiveness comes at substantial computational expense. For example, ERNIE-UniX2~\cite{shan2022ernie} requires 32 A100 GPUs for operation, rendering such approaches impractical for widespread deployment.

Although our proposed PSG framework does not outperform these resource-intensive models, it establishes strong competitiveness within the MMT baseline category. Specifically, on the Multi30K English-German benchmark (Table~\ref{tab:sota_ende}), PSG achieves an average BLEU score of $37.97$, surpassing comparable MMT methods including SAMMT, RG-MMT-EDC, and ConVisPiv by margins of $+1.14$, $+2.84$, and $+2.25$ respectively. As shown in Table~\ref{tab:pt}, when compared to VGAMT~\cite{futeral-etal-2023-tackling} - which utilizes pretrained mBART - PSG exhibits only a modest performance gap of $-1.13$ BLEU points. Notably, while PSG and SAMMT operate at similar model scales, PSG's performance improvement over SAMMT is comparable to the gain achieved by VGAMT despite the latter's significantly larger architecture, further underscoring PSG's efficiency and competitiveness.

Furthermore, our proposed pruning strategy demonstrates model-agnostic characteristics, enabling seamless integration with similar pretrained architectures for potential performance enhancement. This adaptability suggests promising directions for future research in efficient MMT model development.

\subsection{Case Analyses}
Figure~\ref{fig:case} presents the results of some case studies of the PSG model. Overall, the results demonstrate that the PSG model can effectively generate high-quality translations on both En-De and En-Fr translation directions. However, in Case (d), the model fails to interpret the referent of ``another'', while in Case (c), it misidentifies ``gi''. These errors reveal limitations in inter-sentence comprehension and vocabulary, reinforcing the need for pretrained language models as a foundation, which is also a focus of our future research.

%% file: tables/architecture.tex
\begin{table}[t]
    \centering
    \resizebox{1\linewidth}{!}{
    \begin{tabular}{l|c|c|c|c}
        \toprule
        \multirow{2}{*}{\textbf{Model}} & \multicolumn{4}{c}{\textbf{Backbone Size}} \\
        \cmidrule{2-5}
        & Tiny & Small & Medium & Base \\
        \midrule
        \midrule
        \multicolumn{5}{l}
        {\textbf{Architecture}} \\
        \midrule
        Enc./Dec. Layers & $4$ & $6$ & $6$ & $6$\\
        \midrule
        Attn. Heads & $4$ & $8$ & $8$ & $8$\\
        \midrule
        Embedding Dim. & $128$ & $128$ & $256$ & $512$ \\
        \midrule
        \multicolumn{5}{l}
        {\textbf{Optimization}} \\
        
        \midrule
        Dropout & \multicolumn{4}{c}{$0.3$} \\
        \midrule
        Batch Size (Tokens) & \multicolumn{4}{c}{$4,096$} \\
        \midrule
        Warmup Updates & \multicolumn{4}{c}{$20,000$} \\
        \midrule
        Max Updates & \multicolumn{4}{c}{$80,000$} \\
        \midrule
        Learning Rate & $0.0050$ & \multicolumn{2}{c|}{$0.0010$} & $0.0005$ \\
        \bottomrule
    \end{tabular}}
    \caption{Architecture and optimization hyperparameters settings of the PSG variations. (For AmbigCaps, the batch size is increased to $6,400$.)}
    \label{tab:params}
\end{table}

%% file: tables/cost.tex
\begin{table}[t]
    \centering
    \setlength\tabcolsep{4pt}
    \resizebox{1\linewidth}{!}{
    \begin{tabular}{l|c|c|c}
    \toprule
    \textbf{Module}&\textbf{GPU}&\textbf{\#Samples}&\textbf{\#Time(s)}\\
    \midrule
    \midrule
    VSG Generation&$1$ RTX 2080TI&\multirow{5}{*}{$10,000$}&$5,802$\\
    LSG Parsing&$1$ RTX 3090&&$89$\\
    VSG Vectorization&$1$ RTX 3090&&$133$\\
    LSG Vectorization&$1$ RTX 3090&&$297$\\
    MMT w/o prune&$4$ RTX 3090&&$171$\\
    MMT&$4$ RTX 3090&&$177$\\
    
    \bottomrule
    \end{tabular}}
    \caption{The average computational cost of each module in PSG (measured over 10 epochs for MMT, with all results averaged over three runs).}
    \label{tab:cost}
    \end{table}

%% file: figures/bs_ende.tex
\begin{figure}[h]
    \includegraphics[width=1\linewidth]{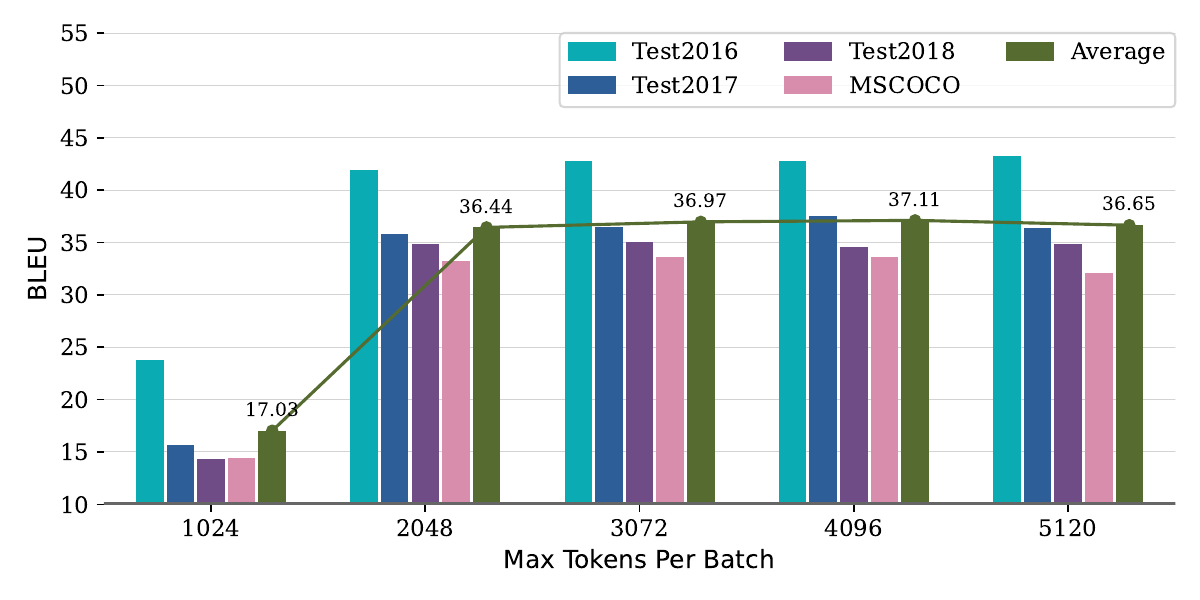}
    \caption{Sensitivity analysis results concerning the training batch size on Multi30K English-German benchmark.}
    \label{fig:bs_ende}
\end{figure}

%% file: figures/bs_enfr.tex
\begin{figure}[h]
    \includegraphics[width=1\linewidth]{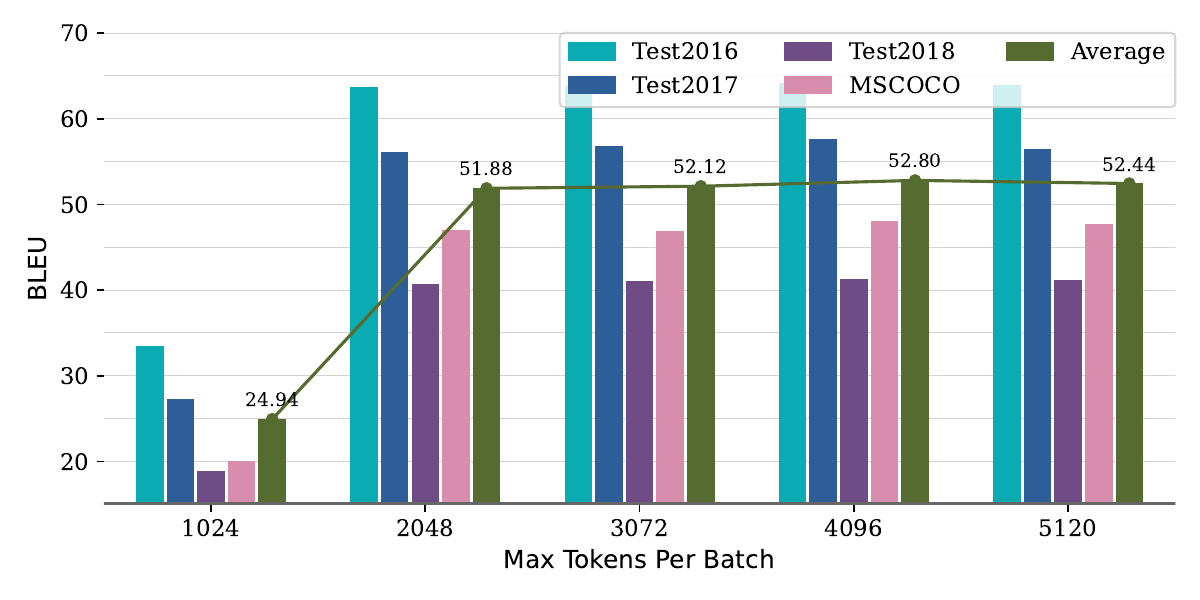}
    \caption{Sensitivity analysis results concerning the training batch size on Multi30K English-French benchmark.}
    \label{fig:bs_enfr}
\end{figure}

%% file: tables/size.tex
\begin{table*}[ht]
    \centering
    \setlength\tabcolsep{4pt}
    \resizebox{1\linewidth}{!}{
    \begin{tabular}{c|ccc|cc|cc|cc|cc|cc}
    \toprule
    \multirow{2}{*}{\textbf{Method}} &\multirow{2}{*}{\textbf{L}} &\multirow{2}{*}{\textbf{H}} &\multirow{2}{*}{\textbf{D}} & \multicolumn{2}{c}{\textbf{Test 2016}} & \multicolumn{2}{c}{\textbf{Test 2017}} &
    \multicolumn{2}{c}{\textbf{Test 2018}} &
    \multicolumn{2}{c}{\textbf{MSCOCO}} & \multicolumn{2}{c}{\textbf{Average}} \\
    \cmidrule(lr){5-14}
     &&&& BLEU & METEOR & BLEU & METEOR & BLEU & METEOR &BLEU & METEOR &BLEU & METEOR \\
    \midrule
    \midrule
    \multicolumn{13}{l}{En$\rightarrow$De}\\
    \midrule
    \multirow{4}{*}{PSG} &$4$& $4$&$128$&$\mathbf{43.37}$ & $\mathbf{69.76}$ & $34.95$ & $63.58$ & $33.56$ & $60.05$ & $31.80$ & $58.52$ & $35.92$ & $62.98$ \\

    & $6$ & $8$ & $128$ & $40.26$& $68.29$ & $32.19$& $61.33$ & $30.50$& $58.24$ & $28.64$ & $56.00$ & $32.89$ & $60.96$\\

    & $6$ & $8$ & $256$ & $42.25$& $68.94$ & $35.29$& $62.81$ & $34.11$& $59.53$ & $33.09$ & $58.44$ & $36.18$ & $62.43$\\

    &\cellcolor{black!6}$6$&\cellcolor{black!6}$8$&\cellcolor{black!6}$512$& \cellcolor{black!6}$42.75$ & \cellcolor{black!6}$69.11$ & \cellcolor{black!6}$\mathbf{37.50}$ & \cellcolor{black!6}$\mathbf{63.87}$ & \cellcolor{black!6}$\mathbf{34.55}$ & \cellcolor{black!6}$\mathbf{59.84}$ & \cellcolor{black!6}$\mathbf{33.66}$ & \cellcolor{black!6}$\mathbf{59.22}$ & \cellcolor{black!6}$\mathbf{37.11}$ & \cellcolor{black!6}$\mathbf{63.21}$ \\

    \midrule
    \multicolumn{13}{l}{En$\rightarrow$Fr}\\
    \midrule

    \multirow{4}{*}{PSG}& $4$ & $4$ &$128$& $63.89$ & $82.24$ & $56.54$ &$77.32$ &$38.18$ & $63.80$ & $47.22$ & $\mathbf{72.02}$ & $51.46$ & $73.85$\\

    & $6$ & $8$ &$128$& $61.88$ & $80.99$& $53.92$ & $75.88$& $37.37$ & $63.18$& $45.97$ & $70.91$ & $49.78$ & $72.74$\\

    & $6$ & $8$ &$256$& $63.83$ & $82.33$& $56.48$ & $77.10$& $39.72$ & $64.43$& $47.18$ & $71.74$ & $51.80$ & $73.90$\\

    &\cellcolor{black!6}$6$&\cellcolor{black!6}$8$&\cellcolor{black!6}$512$& \cellcolor{black!6}$\mathbf{64.22}$ & \cellcolor{black!6}$\mathbf{82.27}$ & \cellcolor{black!6}$\mathbf{57.66}$ & \cellcolor{black!6}$\mathbf{77.73}$ &\cellcolor{black!6}$\mathbf{41.25}$ &\cellcolor{black!6}$\mathbf{65.06}$ & \cellcolor{black!6}$\mathbf{48.06}$ & \cellcolor{black!6}$71.93$ & \cellcolor{black!6}$\mathbf{52.80}$ & \cellcolor{black!6}$\mathbf{74.24}$ \\
    \bottomrule
    \end{tabular}}
    \caption{The impact of Transformer backbones of different sizes on PSG performance on the Multi30K English-German and English-French benchmarks, where L, H, and D represent the number of layers, heads, and dimensions, respectively.}
    \label{tab:size}
    \end{table*}

%% file: tables/pt.tex
\begin{table*}[t]
    \centering
    \setlength\tabcolsep{8pt}
    \resizebox{1\linewidth}{!}{
    \begin{tabular}{l|c|cc|cc|cc|cc}
    \toprule
    \multirow{2}{*}{\textbf{Method}} &\multirow{2}{*}{\textbf{Pretrained Backbone}}&  \multicolumn{2}{c|}{\textbf{Test 2016}} & \multicolumn{2}{c|}{\textbf{Test 2017}} &
    \multicolumn{2}{c|}{\textbf{MSCOCO}} & \multicolumn{2}{c}{\textbf{Average}} \\
    \cmidrule(lr){3-10}
     && BLEU & COMET & BLEU & COMET& BLEU & COMET& BLEU & COMET \\
    \midrule
    \midrule
    \multicolumn{10}{l}{En$\rightarrow$De}\\
    \midrule
    VGAMT~\citeyearpar{futeral-etal-2023-tackling}&mBART&$43.30$&$\mathbf{0.694}$&$38.30$&$0.653$&$35.70$&$\mathbf{0.544}$&$39.10$&$\mathbf{0.630}$\\
    CLIPTrans~\citeyearpar{gupta2023cliptrans}&mBART&$43.87$&$-$&$37.22$&$-$&$34.49$&$-$&$38.53$&$-$\\

    GRAM~\citeyearpar{vijayan-etal-2024-adding}&mBART&$46.50$&$-$&$\mathbf{43.60}$&$-$&$\mathbf{39.10}$&$-$&$\mathbf{43.07}$&$-$\\

    ERNIE-UniX$^2$~\citeyearpar{shan2022ernie}&mBART&$\mathbf{49.30}$&$-$&$-$&$-$&$-$&$-$&$-$&$-$\\
    \rowcolor{black!6}
    PSG&No&$42.75$&$0.351$&$37.50$&$0.256$&$33.66$&$0.169$&$37.97$&$0.259$\\
    \midrule
    \multicolumn{10}{l}{En$\rightarrow$Fr}\\
    \midrule
    VGAMT~\citeyearpar{futeral-etal-2023-tackling}&mBART&$\mathbf{67.20}$&$\mathbf{0.968}$&$\mathbf{61.60}$&$0.921$&$\mathbf{51.10}$&$\mathbf{0.811}$&$\mathbf{59.97}$&$\mathbf{0.900}$\\

    CLIPTrans~\citeyearpar{gupta2023cliptrans}&mBART&$64.55$&$-$&$57.59$&$-$&$48.83$&$-$&$56.99$&$-$\\

    \rowcolor{black!6}
    PSG&No&$64.22$&$0.739$&$57.66$&$0.698$&$48.06$&$0.549$&$56.65$&$0.662$\\
    \bottomrule
    
    \end{tabular}}
    \caption{Comparisons with methods using pretrained backbone on Multi30K.}
    \label{tab:pt}
    \end{table*}

%% file: figures/case.tex
\definecolor{correct}{HTML}{2e8d5d}
\definecolor{incorrect}{HTML}{c84c33}\begin{figure*}[hbt]
    \includegraphics[width=1\linewidth]{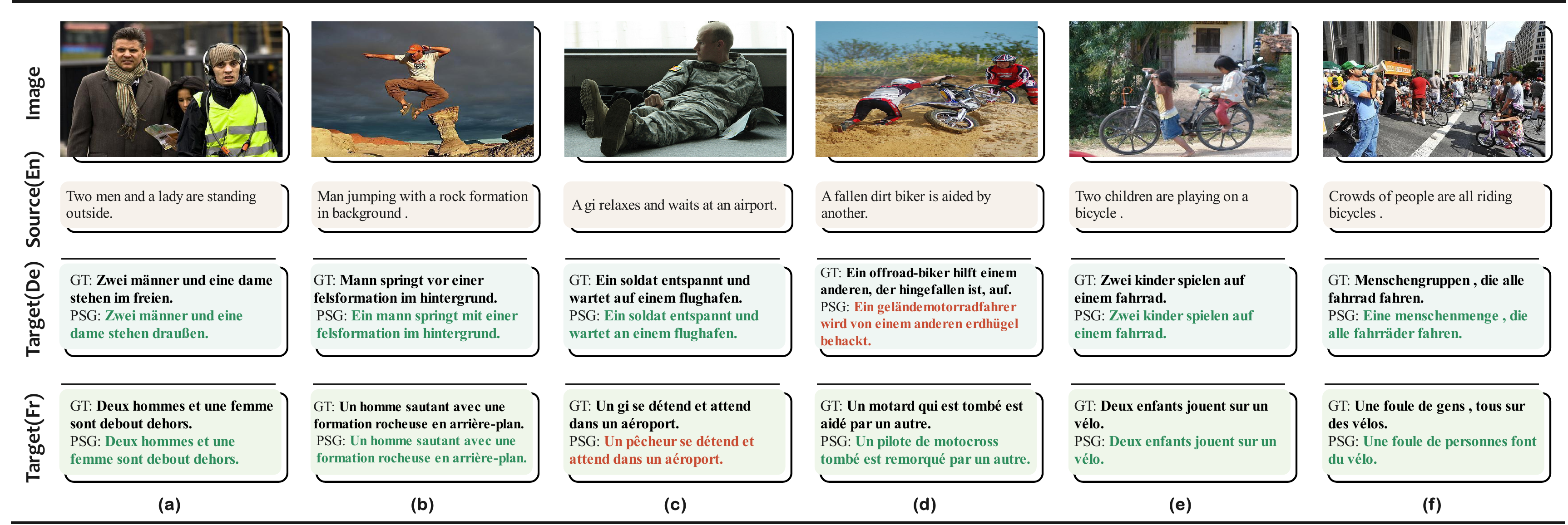}
    \caption{The case study results of our PSG model. Correct predictions are highlighted in \textbf{\textcolor{correct}{green}}, while incorrect ones are marked in \textbf{\textcolor{incorrect}{red}}.}
    \label{fig:case}
  \end{figure*}